\pdfoutput=1

\documentclass[11pt]{article}

\usepackage[]{acl}

\usepackage{times}
\usepackage{latexsym}

\usepackage[T1]{fontenc}
\usepackage[utf8]{inputenc}

\usepackage{microtype}

\usepackage[capitalize,noabbrev]{cleveref}

\usepackage{caption}
\usepackage{subcaption}
\usepackage{enumitem}
\usepackage{rotating}

\usepackage{multirow}
\usepackage{wrapfig,lipsum,booktabs}
\usepackage{etoolbox}

\usepackage{todonotes}
\usepackage{mdframed}
\newcommand{\myparagraph}[1]{\textbf{#1}.}

\title{Do Language Models Know When They're Hallucinating References?}

\author{Ayush Agrawal \\
  Microsoft Research\\
  \texttt{t-agrawalay@microsoft.com} \\\And
  Mirac Suzgun \\
  Stanford University \\
  \texttt{msuzgun@stanford.edu} \\
  \\\AND
  Lester Mackey \\
Microsoft Research \\
  \texttt{lmackey@microsoft.com} \\
  \\\And
  Adam Tauman Kalai \\
  OpenAI\thanks{\ \ Work done while at Microsoft Research.} \\
  \texttt{adam@kal.ai} \\}

\begin{document}
\maketitle

\begin{abstract}
State-of-the-art language models (LMs) are notoriously susceptible to generating hallucinated information. Such inaccurate outputs not only undermine the reliability of these models but also limit their use and raise serious concerns about misinformation and propaganda. In this work, we focus on hallucinated book and article references and present them as the 
``model organism'' of 
language model hallucination research, 
due to their frequent and easy-to-discern nature. We posit that if a language model cites a particular reference in its output, then it should ideally possess sufficient information about its authors and content, among other relevant details. Using this basic insight, we illustrate that one can identify hallucinated references without ever consulting any external resources, by asking a set of \emph{direct} or \emph{indirect} queries to the language model about the references. These queries can be considered as ``consistency checks.'' Our findings highlight that while LMs, including GPT-4, often produce inconsistent author lists 
for hallucinated references,  
they also often accurately recall the authors of real references. 
In this sense, the LM can be said to ``know'' when it is hallucinating references. Furthermore, these findings show how hallucinated references can be dissected
to shed light on their nature.
Replication code and results can be found at \href{https://github.com/microsoft/hallucinated-references}{github.com/microsoft/hallucinated-references}.

\end{abstract}

\section{Introduction}
Despite their unparalleled capabilities, recent large language models (LLMs) still exhibit a tendency to generate seemingly credible yet incorrect or unfounded information. This phenomenon is often referred to as the ``hallucination'' problem in the field of natural-language processing (NLP).\footnote{Though it is an anthropomorphism, we use the term \textit{hallucinate}  due to its widespread adoption, following the use-theory of meaning \citep{Wittgenstein1953-WITPI-4}. Additionally, we use the terms \textit{hallucinate} and  \textit{fabricate} interchangeably throughout the paper.} 
As one might imagine, the ramifications of these hallucination generations can be profoundly detrimental  when these outputs find their way to critical domains such as healthcare, finance, law, or academic publications, where factuality is essential and non-negotiable. In fact, a recent example underlining the gravity of this issue involved a U.S. judge imposing sanctions on two New York lawyers for submitting a legal brief that included several fictitious case citations that were generated by ChatGPT.\footnote{The original newspaper article detailing this incident can be found at this link.~\citep{merken2023}}

The are two primary challenges ahead for both researchers and practitioners within the NLP community. The first requires developing a deeper understanding of why these language models resort to fabricating information, while the second demands creating mechanisms that can not only promptly detect but also mitigate, if not completely prevent, inaccurate information in model outputs. To that effect, in this work, we study the problem of hallucinated book and article references related to the field of computer science and present a simple yet effective method to detect hallucinated references without relying on external tools.

Drawing inspiration from the role of the fruit fly, Drosophila melanogaster, as a model organism in biological research, we suggest that the NLP community focus on the study of hallucinated references to better understand and mitigate wider hallucination challenges. These hallucinated references present distinct characteristics that render them suitable for study. First, their automatic classification is more straightforward than other hallucination varieties.\footnote{In contrast, hallucinations like factoids pose classification challenges due to their nuanced phrasing and the uncertainty regarding their presence in training datasets.} As an illustration, our method that leverages a search engine API closely matches each of four human expert evaluations, in at least 99 out of a sample of 100 references. Moreover, the static nature of academic reference titles, combined with their broad online availability (on platforms like Google Scholar, Semantic Search, and arXiv), suggests they frequently appear in large, popular language modeling corpora. Additionally, many within the research domain already possess the skills and knowledge pertinent to studying these hallucinations. We therefore believe that just as fruit fly studies have enriched our understanding of biology, focusing on these specific reference hallucinations can pave the way for insights and solutions for more complex and challenging hallucination forms.

We outline the rest of this work as follows. 
We are interested in investigating \emph{when and why language models produce hallucinated references} and \emph{what can be done to prevent them}. 
We explore whether LLMs such as GPT-4 can recognize their own hallucinated outputs without relying on any external tools. While this approach does not fully unravel the reasons behind and solutions to hallucinations, it adds valuable perspective. Specifically, if LLMs can identify their own hallucinations, it implies the root of the issue may not lie in training or representation, but rather in the generation (i.e., decoding) process, given that models inherently possess enough data to potentially lower the rate of hallucinations. Our experiments compared different questioning strategies to use the LM to detect its own hallucinations across GPT and Llama based LM's.

\myparagraph{Contributions} There are several contributions of this work. First, we propose the problem of hallucinated computer science references as a model instance worth studying, like Drosophila. Second, we demonstrate that they can be \textit{reliably} and \textit{automatically} classified. Third, we perform a systematic LM study of hallucinated references, enabling us to compare hallucination rates across LMs. Fourth, we introduce \textit{indirect queries} for evaluating hallucinations. Finally, we compare these to \textit{direct queries} across GPT and Llama based LMs. A conclusion of our work for reducing hallucination is the recognition that changing the generation pipeline can certainly help, while it is less clear if training or representation changes are necessary.

\section{Preliminaries and Background}
\label{sec:preliminaries-and-background}
Following \citet{ji_survey_2023}, we define ``hallucination'' as fabricated text that has little or no grounding in the {training data}. It is worth noting that this is sometimes referred to as \textit{open-domain hallucination} to distinguish it from \textit{closed-domain hallucination} \citep[see:][]{ji_survey_2023}.\footnote{Closed-domain hallucination is typically studied in areas like abstractive summarization and machine translation, where the outputs are compared relative to a specific source document to be summarized or translated as opposed to the entirety of the training data.} 
Our usage of the term \emph{hallucination} aligns with the open-domain variant. 
\myparagraph{Distinguishing Groundedness from Correctness}
The measure of \emph{correctness} (or factuality) relies upon a comparison with ground-truth answers. Previous work on hallucination has blurred the line between groundedness and factuality. (Sometimes this distinction is also referred to as \textit{honesty} versus \textit{truthfulness} \citep{evans_truthful_2021}). For example, the common misconception that ``people use 10\% of their brains'' might be considered grounded if it is mentioned in the training data and assumed to be a true statement; however, this does not mean that it is factual, as it is not a scientifically correct statement.

\myparagraph{Evaluating groundedness} 
Perfectly evaluating hallucinations would require access to the LM's training data. An advantage of the hallucinated reference problem is ease of (approximate) evaluation in that exact-match Web search is a reasonable heuristic for groundedness. This is because the vast majority of article titles present in the training data are included in Web search results---articles are meant to be published and shared, and publishers aim to make their work discoverable by search. 
Furthermore, references generally have titles that are specific enough not to spuriously occur on the Web. Regarding other types of hallucinations, besides article names, which cannot be as easily evaluated, we still hope that our methodology and findings would apply, even if evaluating those types of hallucinations would require access to the training data.

\begin{figure*}
    \centering
\includegraphics[width=0.8\textwidth]{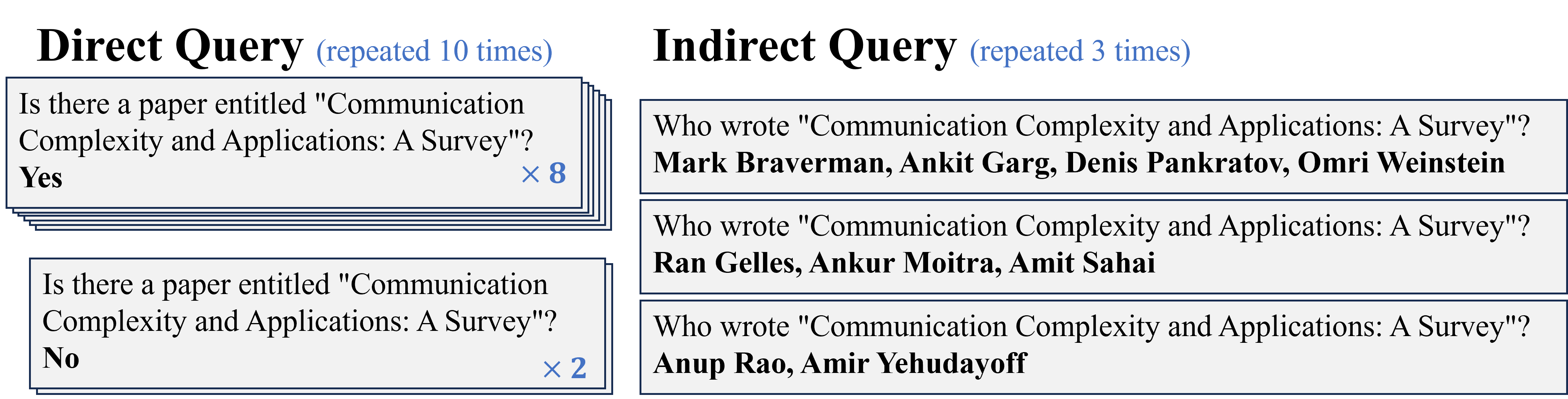}
  \caption{Example direct vs.~indirect LM queries for predicting whether a given paper title is hallucinated or grounded. Direct queries are binary,  repeated multiple times to estimate a probability. Indirect queries are open-ended, and their answers are compared to one another, using the LM, to output an agreement fraction. Language model generations are indicated in \textbf{boldface}. Prompts in this figure have been shortened for illustrative purposes.\label{fig:lead}}
\end{figure*}

\myparagraph{Direct queries (DQs)} 
Our work builds upon and is inspired by two recent works that show how to use black-box generative LMs to assess confidence in generations, without consulting external references or inspecting weights. In particular, \citet{kadavath_language_2022} introduce multiple direct black-box strategies for using an LM to extract confidence estimates by querying the language models on question-answer problems. \citet{manakul_selfcheckgpt_2023} apply a similar direct self-consistency check called SelfCheckGPT to identify relative hallucinations in a summarization context. These queries are direct true/false correctness queries. We test similar approaches in the context of hallucinated references. Black-box generative approaches stand in contrast to the work that either introspects the weights on LMs \citep{azaria_internal_2023}
or that consults existing databases \citep{guo_survey_2022}.

\myparagraph{Indirect queries (IQs)} In addition, we suggest a new approach using what we call \textit{indirect queries}. A direct query may ask, \textit{Is the following paper real?} while an indirect query may ask, \textit{Who are the authors of this paper?}, as illustrated in \cref{fig:lead}. Answers are then generated to the indirect query in $i>1$ independent sessions, and tested for consistency. The motivation for indirect queries comes from investigative interviews, where detectives are advised to interview individuals separately and ask open-ended questions. For instance, consistency may be better evaluated by asking multiple witnesses to \textit{``Describe in detail what the suspect was holding''} rather than asking, \textit{``Was the suspect holding a gun in their right hand?''} \citep{vredeveldt_inconsistent_2014}. In the context of reference hallucination, our hypothesis is that the likelihood of multiple generations agreeing on the same authors for a hallucinated reference would be smaller than the likelihood of multiple responses to a direct query indicating that the reference exists.

\section{Related Work}
\label{sec:related}

Open-domain hallucinations, in the context of GPT-4 discussions \citep{openai_gpt-4_2023, bubeck_sparks_2023}, have garnered attention given their prevalence and associated hazards. \citet[][pg.~82]{bubeck_sparks_2023} comment:
``\textit{Open domain hallucinations pose more difficult challenges, per requiring more extensive research,
including searches and information gathering outside of the session.}''
Yet, our work provides evidence that addressing these hallucinations can be achieved without turning to external resources.

As mentioned, there are multiple definitions of hallucination. In this work, we use the term hallucinations to mean fabricated text that is not grounded in the training data. Factually incorrect generations can be decomposed into two types of errors \citep{evans_truthful_2021}: grounded errors which may be due to fallacies in the training data (e.g., that people use only 10\% of their brains) and ungrounded errors. These two types of errors may need different techniques for remedy. The grounded errors may be reduced by curating a training set with fewer errors or other techniques such as RLHF \citep{ouyang_training_2022}. However, the ungrounded errors which we study\footnote{One can also imagine ungrounded correct generations, such as a generated paper title that exists but is not in the training data, but we find these to be quite rare.} are a fascinating curiosity which still challenge the AI community and one which is not clearly addressable by improving training data.  

There is comparatively little prior work studying \textit{open-domain groundedness} like ours. Some work \citep[e.g.,][]{guu_simfluence_2023} in attribution aims to understand which training examples are most influential in a given output. In recent independent work in the health space, \citet{athaluri_exploring_2023} did an empirical evaluation of hallucinated references within the medical domain. Similar to our approach, they used a Google search for exact string match as a heuristic for evaluating hallucinations. Our study of hallucinated references enables us to estimate the hallucination rates of different models, and, as discussed in prior work, the hallucination problem interestingly becomes more pressing as models become more accurate because users trust them more \citep{openai_gpt-4_2023}.

Related recent works include black-box techniques for measuring confidence in LM generations. Although these works are targeted at factual confidence, the approaches are highly related to our work. While \citet{kadavath_language_2022} use probability estimates drawn from LMs, it is straightforward to extend their procedures to generation-only LMs like ChatGPT using sampling. \citet{lin_teaching_2022} show that LMs can be used to articulate estimates by generating numbers or words as we do. Finally, \citet{manakul_selfcheckgpt_2023} perform self-checks in the context of summarizing a document. All of these works use direct queries which influenced the design of our direct queries. 

Due to space limitations, we do not discuss the work studying closed-domain hallucination (e.g., in translation or summarization) but instead refer the reader to recent survey of \citet{ji_survey_2023}.

\section{Methodology: Consistency Checks}
\label{sec:methodology}

We now provide an overview of our simple yet effective consistency check methodology, explaining how we perform a series of \emph{direct} and \emph{indirect} queries to detect hallucinated references.\footnote{Note that this pipeline is run separately for each of our LMs, so there is no mixing across LMs.} 

\subsection{\emph{Direct} Queries}
\label{sec:direct}
\begin{figure*}[ht]
    \centering
    \includegraphics[width=0.8\textwidth]{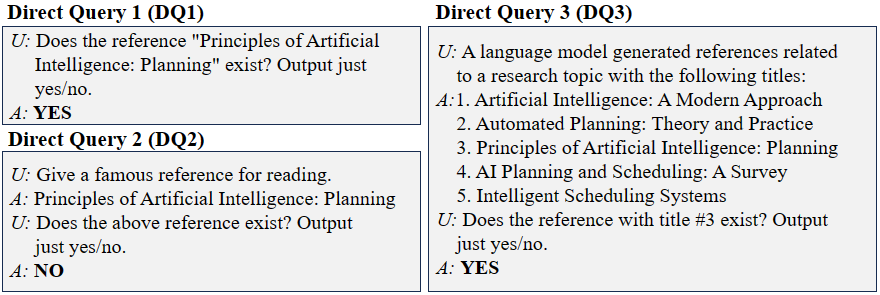}
    \caption{Examples of the three direct prompt templates used for the direct queries, instantiated with candidate reference titles.\label{fig:direct}} 
\end{figure*}

The direct query (DQ) method examines if a particular title exists using a format illustrated in \cref{fig:direct}. We use three simple DQ templates (DQ1, DQ2, and DQ3), drawing insights from \citet{kadavath_language_2022,manakul_selfcheckgpt_2023}. In each case, an LM to expected to answer ``yes'' if it believes that the reference \emph{actually} exists and ``no'' otherwise.  

DQ1 asks outright if the reference does indeed exist. While being simple, this approach can sometimes be problematic as some chat-bot-based LMs have strong biases in answering questions when phrased in a particular way (without any proper context)~\citep{prompt_sensitivity_1}.
DQ2 and DQ3, on the other hand, incorporate context by stating that the reference was generated by an LM or an assistant. Moreover, DQ3 takes it a step further by providing additional references  for comparison, an approach advocated in \citet{kadavath_language_2022}. 

For each query, we generate $j \ge 1$ completions to approximate the probability distribution of the model about the existence of the generated reference.\footnote{For both direct and indirect queries, we employ a temperature rate of $1$ when $j>1$ (i.e., generating multiple completions) and $0$ when $j=1$ (i.e., generating a single completion). The choice of $0$ is intended to capture the model's top pick if a single output is generated.} We measure the \emph{groundedness} rate (see~\cref{sec:preliminaries-and-background}) by dividing the number of completions containing the word ``yes'' by the total number of completions.\footnote{This means that empty or otherwise invalid answers are assigned ``no.'' We do not assume that this score is calibrated as our analysis considers arbitrary probability thresholds.}
We also consider an \emph{ensemble direct query}, denoted by DQ, that simply averages the scores of DQ1, DQ2, and DQ3.

\subsection{\emph{Indirect} Queries}\label{sec:indirect}

The indirect query (IQ) method involves two main steps: \emph{interrogation} and \emph{overlap estimation}.

\begin{figure*}[ht]
    \centering
    \includegraphics[width=0.8\textwidth]{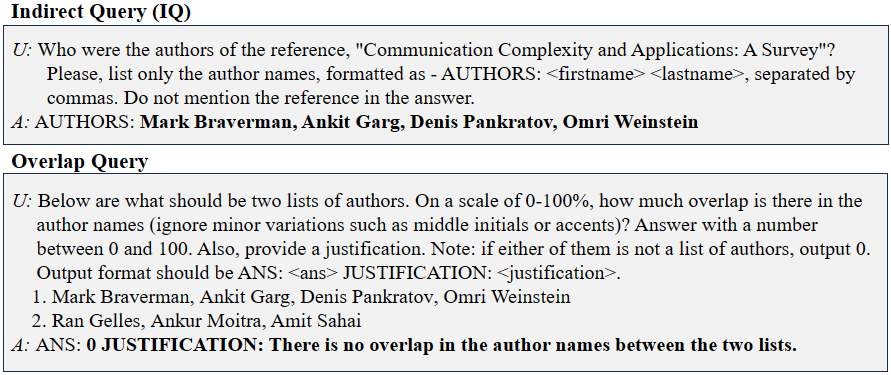}
    \caption{Top: Example of the Indirect Query prompt templates instantiated with a candidate title. Bottom: An example of how we estimate overlap between a pair of answers using the LM.\label{fig:indirect}}
\end{figure*}

 \myparagraph{Step 1: Interrogation} For each reference, we first pose $j$ indirect queries to the LM, asking about the authors of the generated reference, for instance, as shown in \cref{fig:indirect} (top).

\myparagraph{Step 2: Overlap estimation.} Next, we asses the degree of similarity (overlap) between the model responses from the previous step by using a separate query template, as shown in \cref{fig:indirect} (bottom). We initially tested string-matching techniques which we found to be inaccurate and required hyperparameters. Name matching is known to be a thorny problem and one which we found could be performed accurately when using pretrained LMs to compare in pairs.\footnote{It is worth noting that LMs sometimes return responses that do not consist of a list of authors (e.g., a long response beginning with \textit{``I could not find a specific reference titled...''}. In such cases, we simply set the overlap rate to $0$. We also note that traditional parsing and string-matching techniques could be leveraged as an alternative to LMs in this overlap estimation phase.}

The intuition behind our approach is simple: If a language model provides similar (that is, consistent) responses to multiple indirect queries, it can then be assumed that the model is most likely familiar with the reference and that it has seen the reference during its training; such a reference could therefore be deemed \emph{grounded}. 
On the other hand, varied responses might signal that the model does not intrinsically possess knowledge about the author(s) and content of the reference; hence, it can be speculated that the model has presumably not seen the reference during its training and that the reference is mostly likely fabricated.

We also consider an ensemble IQ+DQ check that averages the scores of IQ and the DQ ensemble. 

Finally, we highlight that our consistency checking methods do not rely on external resources such as Google Scholar or Semantic Search. It instead uses the same language model throughout the hallucination detection process.

\begin{figure*}
    \centering
    \includegraphics[width=0.8\textwidth]{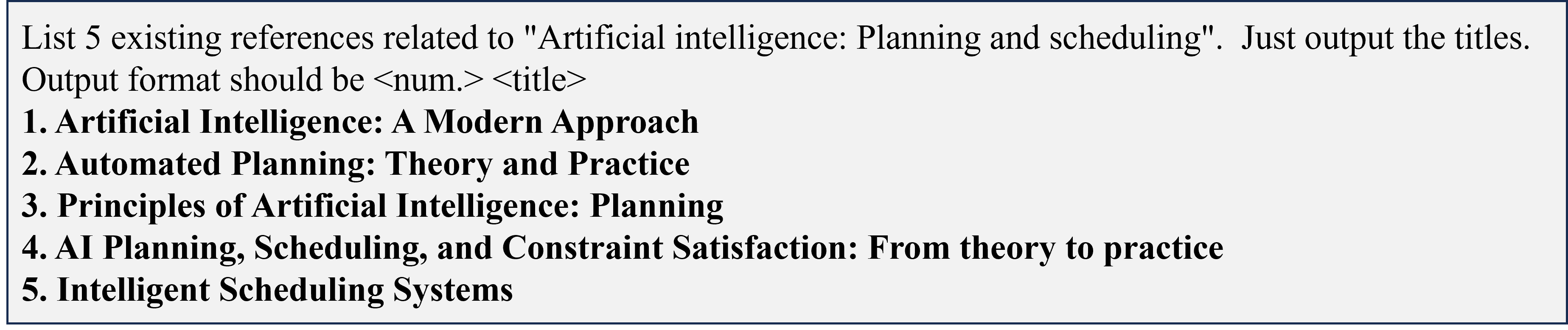} 
    \caption{The prompt used to generate 5 reference titles. This method generates both grounded and hallucinated references. Topics are chosen from the ACM Computing Classification System.}\label{fig:gen}
\end{figure*}

\section{Experimental Details}
Here, we describe the steps taken to build a corpus of article and book references pertaining to computer science topics for each language model, as well as the automatic labeling heuristic used to annotate these generated references. 

\subsection{Dataset Construction Using ACM CCS}
To ensure that our corpus of references is representative of a broad spectrum of the topics in computer science, we used the \href{https://dl.acm.org/ccs}{ACM Computing Classification System}~\citep[CCS;][]{rous_major_2012} as our main source. The CCS provides a structured taxonomy for computer science, ranging from 12 high-level subjects down to 543 specific topics. 

From the 543 topics, we selected a uniformly random subset of 200 topics, each denoted as \textit{area: topic} (e.g., \textit{Information retrieval: Retrieval models and ranking}). For each chosen topic, we then prompted each LM to generate five related  reference titles, amounting to 1,000 total titles per LM as shown in~\cref{fig:gen}.

\subsection{Automatic Labeling and Verification}\label{sec:automated}
Next, we employed the \href{https://www.microsoft.com/en-us/bing/apis/bing-web-search-api}{Bing search engine API}\footnote{\url{https://www.microsoft.com/en-us/bing/apis/bing-web-search-api}}
as an automatic labeling heuristic, labeling each of the 1,000 reference titles generated in the previous step as either \emph{grounded} (G) or \emph{hallucinated} (H) based on exact matches. The reference title surrounded by quotes is searched in the web (e.g., “LMs are few-shot learners”). We label the reference as hallucinated if no results are retrieved and as grounded otherwise.

To assess the efficacy of this automated pipeline, we asked four expert annotators (all computer scientists familiar with academic writing and publication) to manually label 10\% of the GPT-4-generated references. 
One of the annotators agreed with Bing on 100\% of the labels, and the other three each had 99\% agreement with Bing, indicating strong support for the reliability of the automatic labeling pipeline.
See \cref{sec:bing-reliability} for more details. 

\subsection{Models and Parameters}

We evaluate the OpenAI LMs GPT-3 (\textit{text-davinci-003}), ChatGPT (\textit{gpt-35-turbo}), and GPT-4 (\textit{gpt-4}) using the \hyperlink{https://azure.microsoft.com/en-us/products/cognitive-services/openai-service}{Azure OpenAI API} and the open-source  Llama 2 Chat \textit{llama-2-*-chat} series LMs abbreviated as L2-7B, L2-13B, and L2-70B \citep{llama2}. %

We select $i=3$ indirect query results and take the average of the overlapping evaluations to compute the final score for each indirect query experiment. For direct query experiments, we sample $j=10$ judgments at temperature 1.0 and report the fraction of \textit{yes} responses as a final score. 

\subsection{Metrics}

\textbf{Receiver Operating Characteristic (ROC) Curves.} Since each of our querying strategies outputs a real-valued score, one can trade off accuracy on G (i.e., how often truly grounded references are labeled G) and H (how often truly hallucinated references are labeled H) by thresholding the score to form a G or H classification. We visualize this trade-off using a standard receiver operating characteristic (ROC) curve~\citep{roc} and summarize overall detection performance using the area under the ROC curve (AUC).

\textbf{False Discovery Rate (FDR) Curves.} Each groundedness classifier can also be used as a filter to generate a list of likely grounded references for a literature review based on the raw generations of an LM.  Aside from relevance, which we do not study in this work, two primary quantities of interest to a user of this filter would be the fraction of references preserved (more references provide a more comprehensive review) and the fraction of preserved references which are actually hallucinations. We show how these two quantities can be traded off using false discovery rate (FDR) curves. As one varies the threshold of G/H classification and returns only those references classified as grounded, the FDR captures the fraction of references produced which are hallucinations. Users may have a certain rate of tolerance for hallucinations, and one would like to maximize the number of generated references subject to that constraint.

\section{Results and Discussion}\label{sec:experiments}
In this section, we discuss the performance of the indirect and direct methods using  quantitative metrics, and present interesting qualitative findings.

\subsection{Quantitative Analysis}\label{sec:quantitative}
\cref{tab:hallucination-rates} shows the rates of hallucination for the six models studied. As expected, references produced by the newer models (which achieve higher scores on other benchmarks \citep{bigbench}) also exhibit a higher grounding rate or, equivalently, a lower hallucination rate.

\begin{table}[ht]
\resizebox{7.5cm}{!}{
\begin{tabular}{l|llllll}
\toprule
 \textbf{LLM} & \textbf{GPT-4} & \textbf{ChatGPT} & \textbf{GPT-3} & \textbf{L2-70B} & \textbf{L2-13B} & \textbf{L2-7B} \\
\midrule
\textbf{H\%} & 46.8\% & 59.6\% & 73.6\% & 66.2\% & 76.7\% & 68.3\% \\
\bottomrule
\end{tabular}}
\caption{The hallucination rate (out of 1000 generated titles), as determined by ground-truth labels assigned using the Bing search API.\label{tab:hallucination-rates}}
\end{table}

Due to space limitations, we show the ROC and FDR curves for GPT-4, ChatGPT, and L2-70B and defer additional LM results to \cref{sec:additional-experiments}. 

\begin{figure*}[ht]
     \centering
          \begin{subfigure}{\textwidth}
         \centering
    \includegraphics[width=\textwidth]{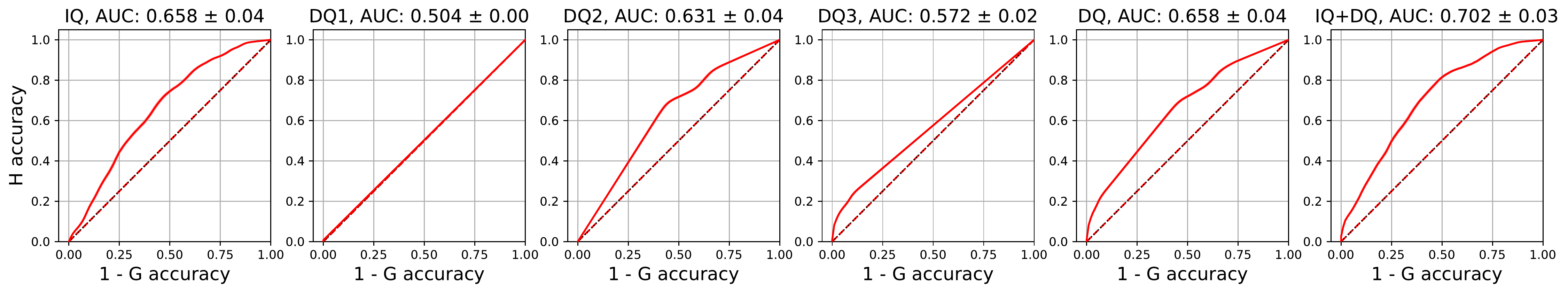}
  \caption{Llama-2-70b-chat}
  \label{fig:llama70_roc}
     \end{subfigure}
     \hfill
     \begin{subfigure}[b]{\textwidth}
         \centering
\includegraphics[width=\textwidth]{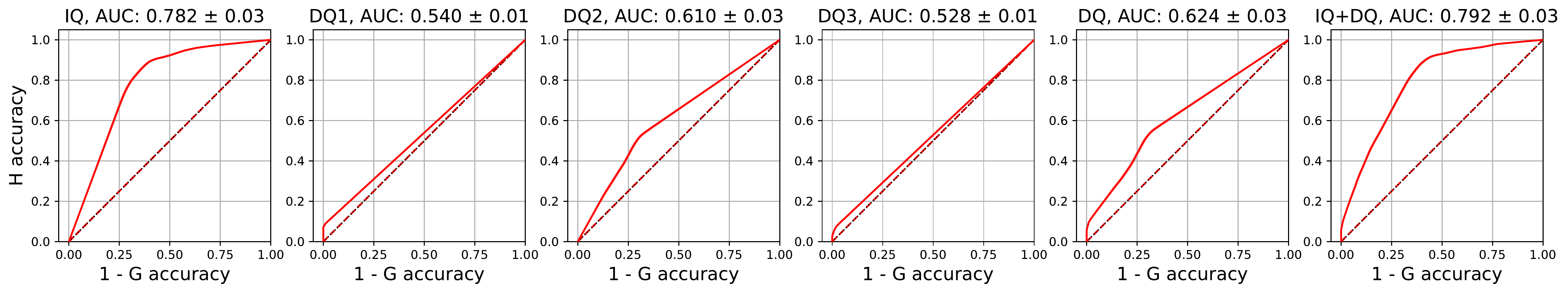}
   \caption{ChatGPT}
  \label{fig:chat_gpt_roc}
     \end{subfigure}
     \hfill
     \begin{subfigure}[b]{\textwidth}
         \centering
    \includegraphics[width=\textwidth]{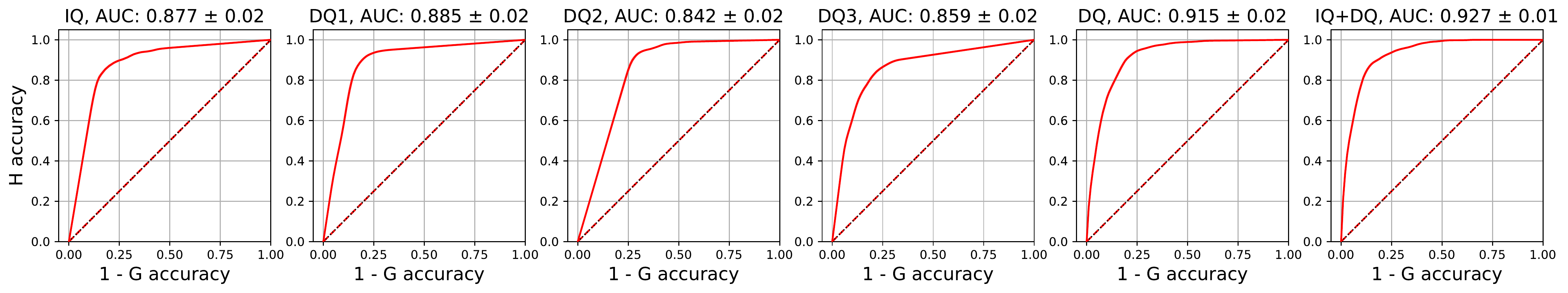}
\caption{GPT-4}
  \label{fig:gpt4_roc}
     \end{subfigure}
        \caption{%
        For each individual (IQ, DQ1-3) and ensemble (DQ, IQ+DQ) consistency check, we display the trade-off between accuracy on grounded and hallucinated references with
        95\% confidence bands  based on 100 bootstrap replicates and a 95\% confidence interval for the AUC using the \citet{delong1988comparing} estimate of standard error.
        \label{fig:three_roc_graphs}\vspace{-4pt}}
\end{figure*}

The ROC curves are shown for each approach and model in \cref{fig:three_roc_graphs}. These figures enable one to explore different points on this trade off for each classifier. For the L2-70B and ChatGPT models, the IQ procedure performs best overall as  quantified via AUC. For GPT-4 (\cref{fig:gpt4_roc}), both the IQ and DQ approaches work well for classifying hallucination and groundedness with the IQ (AUC: $0.878$)  and DQ1 (AUC: $0.887$) performing the best. The performance of each  procedure generally improves as the model size increases.

\begin{figure*}[ht!]
\centering  
\newcommand{\widthscale}{0.30}  
\centering  
\includegraphics[width=\widthscale\textwidth]
{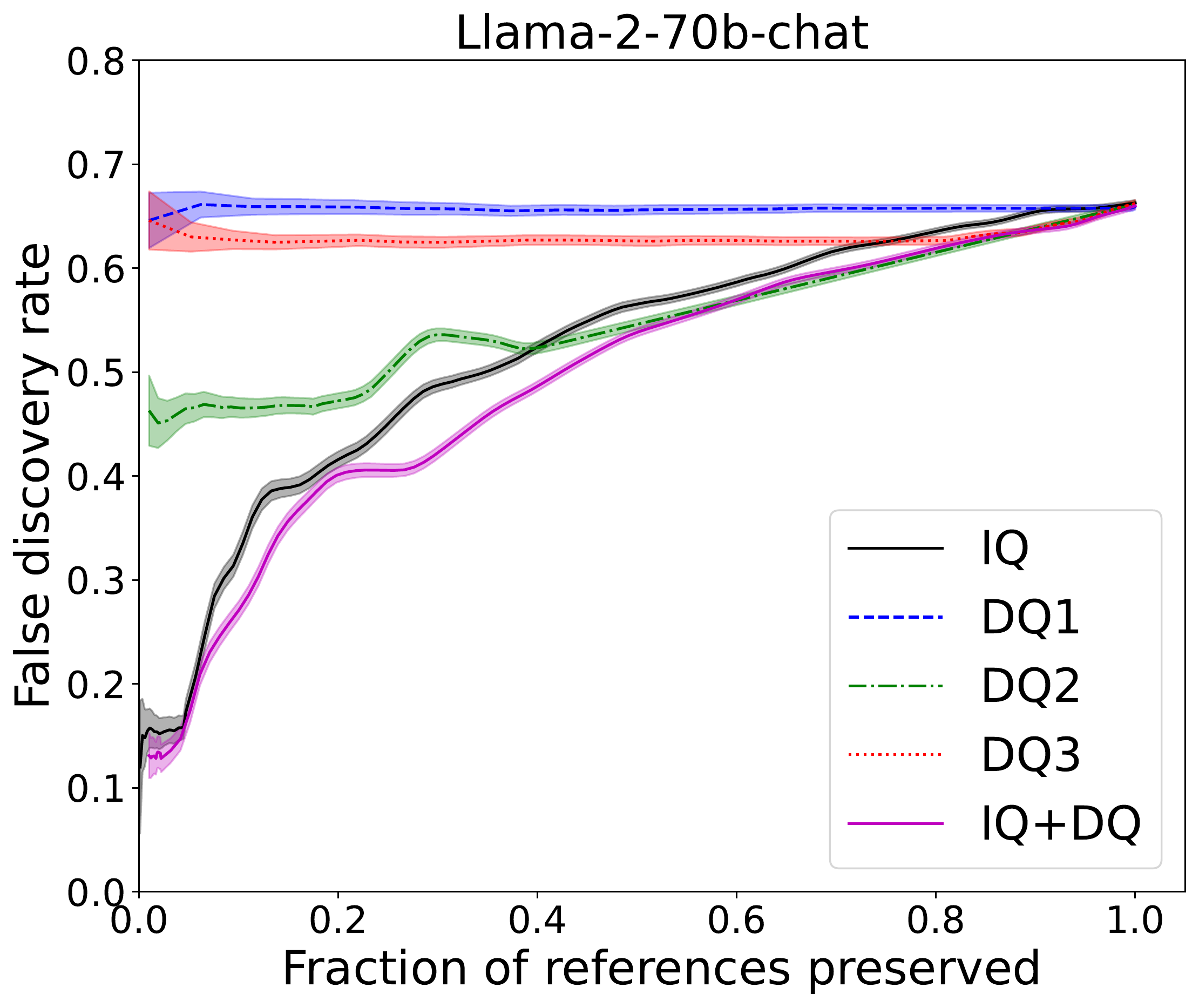}
\includegraphics[width=\widthscale\textwidth]
{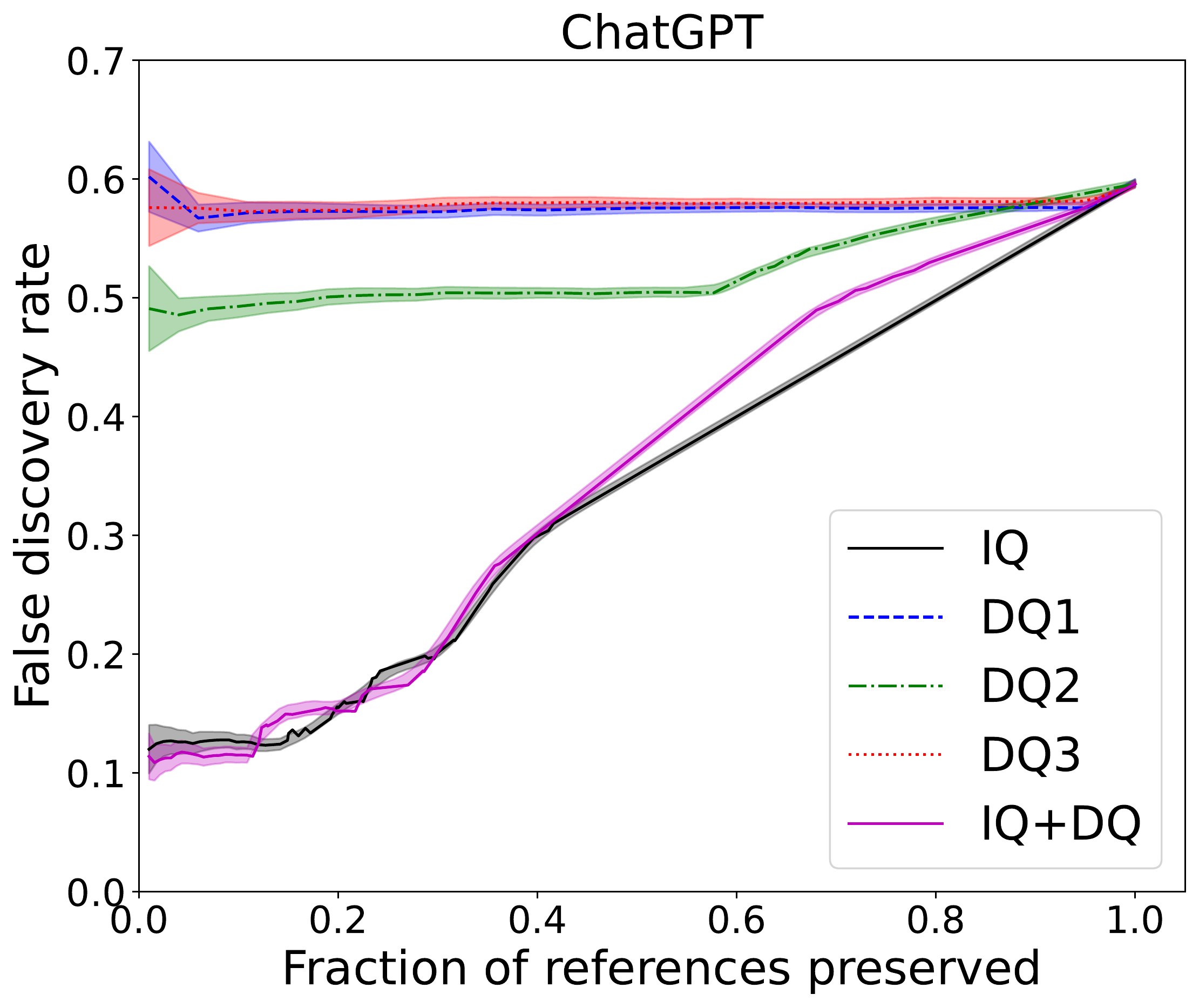}
\includegraphics[width=\widthscale\textwidth]
{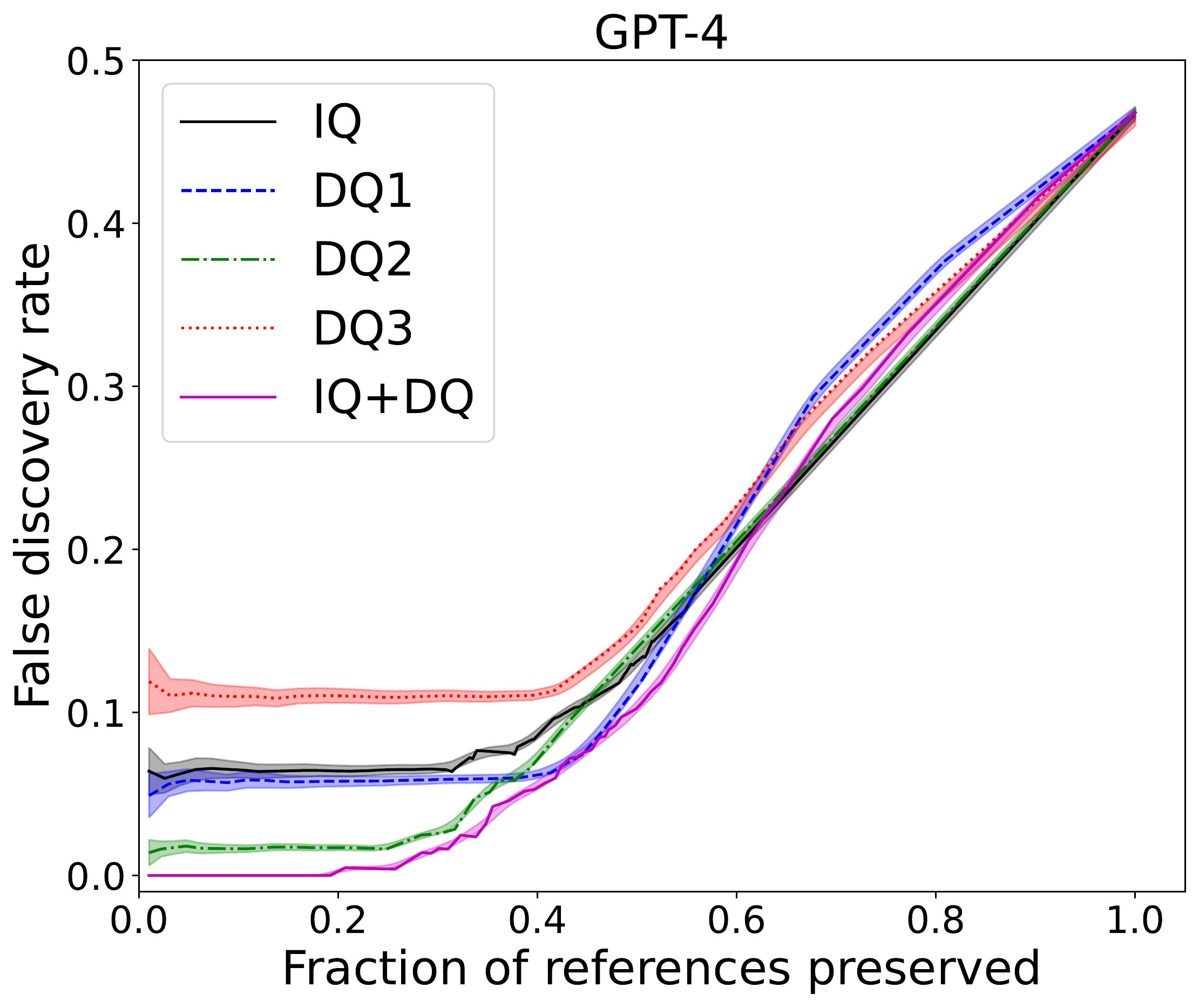}
\caption{%
False discovery rate (FDR) vs.\  fraction of references preserved for each groundedness filter and LM. 
We compute 95\% confidence intervals from a 100-replicate bootstrap mean $\pm1.96$ times the bootstrap standard
error.
\label{fig:fdr_curves_cf}
}
\end{figure*}

\cref{fig:fdr_curves_cf} shows FDR curves for the three models. For L2-70B and ChatGPT, the IQ method achieves significantly lower FDR and a provides a substantially better FDR-preservation rate trade-off than the other approaches.  For GPT-4, both IQ and DQ methods offer low FDR with comparable trade-offs.

Overall, IQ appears to be more accurate than DQ1-3 for ChatGPT and  L2-70B, while for GPT-4 DQ1-3 and IQ were similarly effective. 
For each LM, ensembling further boosts classification performance with the IQ+DQ ensemble obtaining the best AUC and lower FDR curves for each LM.

The compute costs, which involve $\approx$6.6 million tokens and \$412, are discussed in Section~\ref{sec:compute}.

\subsection{Qualitative Findings}
A qualitative examination of the titles generated by the LMs and their classifications according to the Bing search API revealed several interesting observations:
1) \textit{Title mashups:} Many hallucinated titles were combinations of multiple existing titles. For example, a hallucinated title ``Privacy-Preserving Attribute-Based Access Control in Cloud Computing" could be ``fabricated" from (of the many possibilities) existing titles ``Privacy-Preserving Attribute-Based Access Control for Grid Computing" and ``Access Control in Cloud Computing".
2). \textit{Bing's search flexibility:} The Bing quoted search heuristic is more lenient than exact match, ignoring more than just capitalization and punctuation. However, presumably since Bing quoted search is designed to facilitate title searches, it works well. 3) \textit{Deceptive plausibility:} Some hallucinations were ``plausible sounding'' such as \textit{A survey on X} for topic \textit{X}, even when such a survey did not exist. 4) \textit{DQ's false positives:}  \label{false-positives} Direct methods may fail to identify hallucinations on ``plausible sounding'' titles such as surveys or book chapters. The indirect method also sometimes failed to identify a hallucination because the LM would consistently produce a ``likely author'' based on the title, for a given non-existent paper. For example, GPT-4  hallucinated the title \textit{Introduction to Operations Research and Decision Making}, but there is a real book called \textit{Introduction to Operations Research}. In all three indirect queries, it hallucinated the authors of the existing book, \textit{Hillier Frederick S., Lieberman Gerald J.}. Similarly, for the hallucinated title \textit{Exploratory Data Analysis and the Role of Visualization},  2 of 3 indirect queries produced \textit{John W. Tukey}, the author of the classic, \textit{Exploratory Data Analysis}.
5) \textit{IQ's false negatives:} \label{false-negatives} The indirect method may sometimes fail to identify a grounded paper title which it can recognize/generate, as it may simply not be able to generate authors not encoded in its weights. %
Since, in many applications, identifying potential hallucinations is more important than recognizing all grounded citations, errors due to falsely marking an H as a G are arguably more problematic than classifying a G as an H. A manual examination of 120 examples is given in~\cref{sec:examples}.
\section{Conclusions}
\label{sec:conclusions}
Open-domain hallucination is an important but slippery concept that is difficult to measure. By studying it in the context of references using search engine results, we can quantitatively compare hallucinations across LMs and we can also quantitatively compare different black-box detection methods. Of course, for the sole purpose of detection, one could achieve higher accuracy by directly consulting curated publication indexes. However, we hope that our study of black-box self-detection of hallucinated references sheds light on the nature of open-domain hallucination more broadly, where detecting hallucinations is more challenging. It suggests that hallucination is not entirely a problem of training but rather one that can be addressed using only the same internal model representation with different generation procedures. While our direct and indirect query methods are only partially reliable and impractically expensive, we hope they may pave the way towards more efficient methods that generate text with fewer hallucinations and thereby reduce potential harms of language models. 

There are several directions for future work. 1) \textit{Improved decoding techniques:} An important consequence of our work is the recognition that reducing hallucination may be a problem at generation time. Thus, inventing improved (non-black-box) generation procedures is thus a crucial direction for future work. 2) \textit{Additional indirect questions:} One may improve accuracy by adding more indirect questions such as year or venue. These pose additional challenges as a paper with the same title and authors may often appear in multiple venues (e.g., arXiv, a workshop, a conference, and a journal) in different years. 3) \textit{Generalisability:} It would be very interesting to see if the methods we employ could be used to identify other types of open-domain hallucinations beyond references. Even though hallucinated references are often given as a blatant example of hallucination, perhaps due to the ease with which they can be debunked, these other types of hallucination are also important. Following the investigative interviewing analogy, one way to aim to discover general hallucinations would be to query the LM for ``notable, distinguishing details'' about the item in question. One could then use an LM to estimate the consistency between multiple answers. However, as mentioned for other domains besides references, it may be impossible to determine whether or not a generation is a hallucination without access to the training set (and unclear even with such access).

\section{Limitations}
\label{sec:limitations}

There are several limitations of this work: 1) \textit{Inaccessible training data:} We consider web as a contending proxy for the models' training data. However, we cannot conclude what is truly grounded versus hallucination since we do not have access to the training data. 2) \textit{Hallucination spectrum:} The notion of hallucination is not entirely black and white as considered in this work and in prior works. For example, a generated reference that is a substring or superstring of an existing title is hard to classify with the binary scheme. 3) \textit{Prompt sensitivity:} LMs are notoriously sensitive to prompt wording \citep{prompt_sensitivity_1,prompt_sensitivity_2,prompt_sensitivity_3,prompt_sensitivity_4}. Thus, some of our findings comparing direct and indirect queries may be sensitive to the specific wording in the prompt. 4) \textit{Domain-specific reference bias:} Since we use ACM Computing Classification System for our topics, the results are biased towards computer science references, though it would be straightforward to re-run the procedure on any given list of topics. 5) \textit{Gender and racial biases:} LMs have been shown to exhibit gender and racial biases \citep{swinger_what_2019} which may be reflected in our procedure--in particular: our procedure may not recognize certain names as likely authors, or it may perform worse at matching names of people in certain racial groups where there is less variability in names. Since our work compares LMs and hallucination estimation procedures, the risk is lower compared to a system that might be deployed using our procedures to reduce hallucination. Before deploying any such system, one should perform a more thorough examination of potential biases against sensitive groups and accuracy across different research areas.

\bibliography{references,references_zotero}

\appendix

\section{Bing Search Reliability}
\label{sec:bing-reliability}
\begin{figure*}[h] 
    \centering  
    \includegraphics[width=0.8\textwidth]{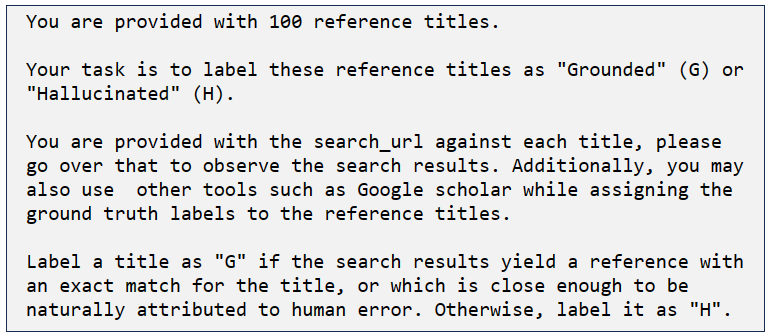}  
    \caption{Labeling instructions shown to the expert human annotators.}  
    \label{fig:ann_ins}  
  \end{figure*}

\begin{figure*}[ht!]
    \centering
    \includegraphics[width=0.8\textwidth]{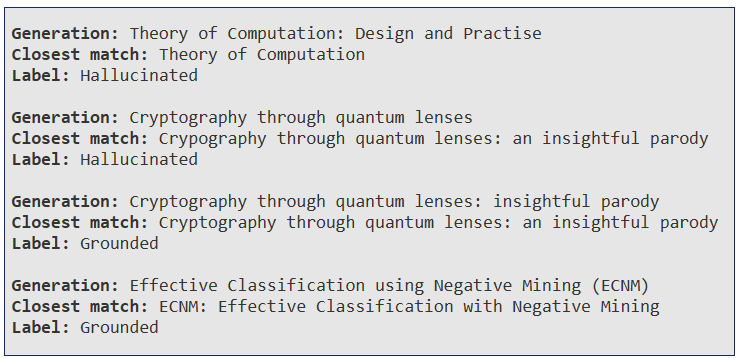}
    \caption{Exemplar labels upon which all expert human annotators agreed prior to assigning manual labels. \label{fig:manual-egs}}
    
\end{figure*}

 Before assigning manual grounded or hallucination labels to each reference title, each expert annotator was given the instructions  shown in~\cref{fig:ann_ins}. Along with a given reference title, the annotators were provided with a  corresponding Google search link as shown in~\cref{tab:ann_sample}. For consistency, the human labelers also agreed on the labels for the four exemplars shown in~\cref{fig:manual-egs}.

We show inter-rater reliability agreement computed using Cohen's $\kappa$ score \citep{kappa} between the labelers and the automated Bing labels in \cref{tab:kappa}. The results demonstrate that the automated labeling generated via Bing search exact match reliably matches the judgments of human experts.

  \begin{table*}[h]
    \centering  
    \small
    \caption{Sample of 2 titles out of 100 titles given to the expert human annotators for labeling.}  
    \label{tab:ann_sample}  
    \begin{tabular}{ccc}  
      \toprule
      \textbf{Reference Title} & \textbf{Search Url} & \textbf{(H/G)}\\ \midrule
      Introduction to Autonomous Robots: Mechanisms, Sensors, Actuators, and Algorithms    & \href{https://www.google.com/search?q=\%22Introduction+to+Autonomous+Robots:+Mechanisms,+Sensors,+Actuators,+and+Algorithms\%22&tbs=cdr:1,cd_max:09/30/2021}{link}
  & ?   \\
  \midrule
      Timing Aware Placement and Routing in FPGAs    & \href{https://www.google.com/search?q=\%22Timing+Aware+Placement+and+Routing+in+FPGAs\%22&tbs=cdr:1,cd_max:09/30/2021}{link}  & ?   \\ 
      \bottomrule
    \end{tabular}  
  \end{table*}  
 
\begin{table}[ht]  
\centering  
\caption{Cohen's $\kappa$ measure of inter-rater reliability between each pair of expert human evaluators and between each expert and the automated Bing labeling described in \cref{sec:automated}. The range of Cohen's $\kappa$ is $[-1,1]$ with a value of $1$ indicating perfect agreement. A value above $0.9$ is considered ''almost perfect'' agreement~\citep{kappa}.}  
\label{tab:kappa}  
\begin{tabular}{lc}  
\toprule  
 & \textbf{Cohen's kappa ($\kappa$)} \\  
\midrule  
\textbf{person A and person B}     & 0.96 \\  
\textbf{person A and person C}  & 0.98\\  
\textbf{person B and person C} & 0.98 \\ 
\textbf{person D and person A} & 0.96 \\
\textbf{person D and person B} & 1.0 \\
\textbf{person D and person C} & 0.98 \\
\textbf{person A and Bing} & 0.98\\  
\textbf{person B and Bing} & 0.98 \\  
\textbf{person C and Bing} & 1.0 \\  
\textbf{person D and Bing} & 0.98 \\
\bottomrule  
\end{tabular}  
\end{table}  

\section{Supplementary Experimental Details}
\label{sec:additional-experiments}

\begin{figure*}[ht]
     \centering
          \begin{subfigure}{\textwidth}
         \centering
    \includegraphics[width=\textwidth]{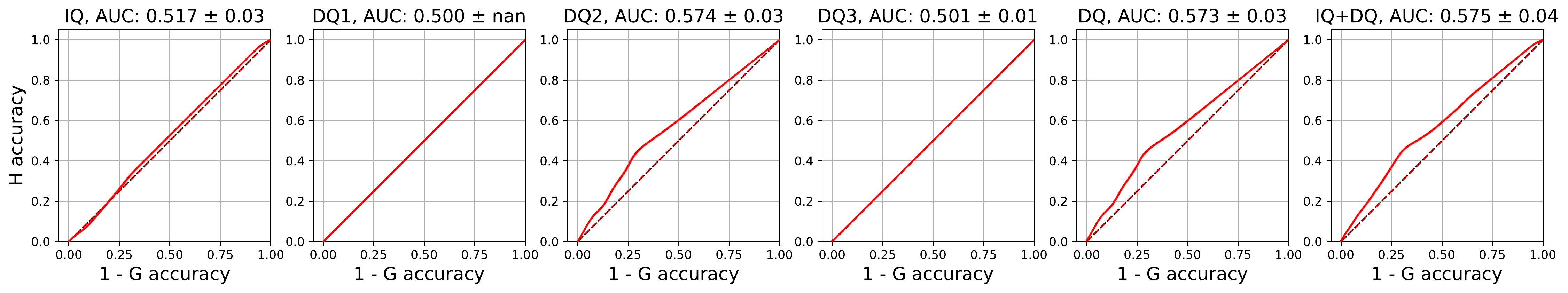}
  \caption{Llama-2-7b-chat}
  \label{fig:llama7_roc}
     \end{subfigure}
     \hfill
     \begin{subfigure}[b]{\textwidth}
         \centering
\includegraphics[width=\textwidth]{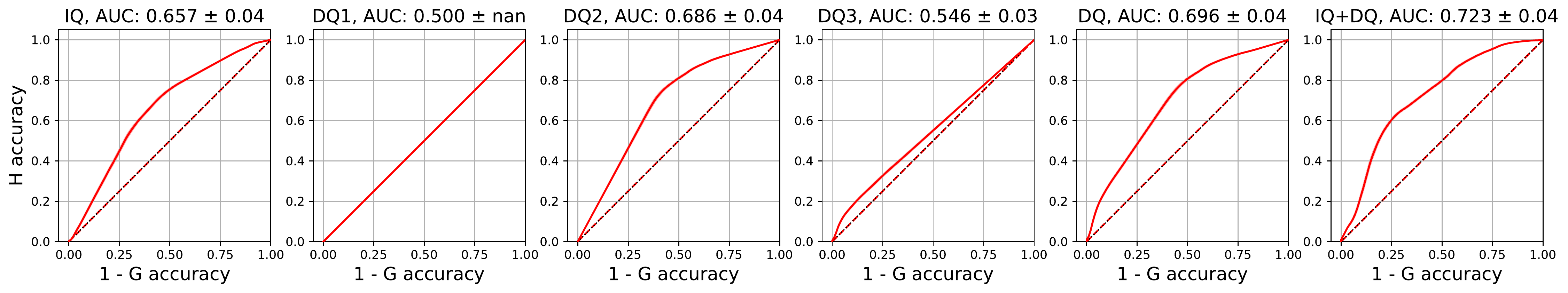}
   \caption{Llama-2-13b-chat}
  \label{fig:llama13_roc}
     \end{subfigure}
     \hfill
     \begin{subfigure}[b]{\textwidth}
         \centering
    \includegraphics[width=\textwidth]{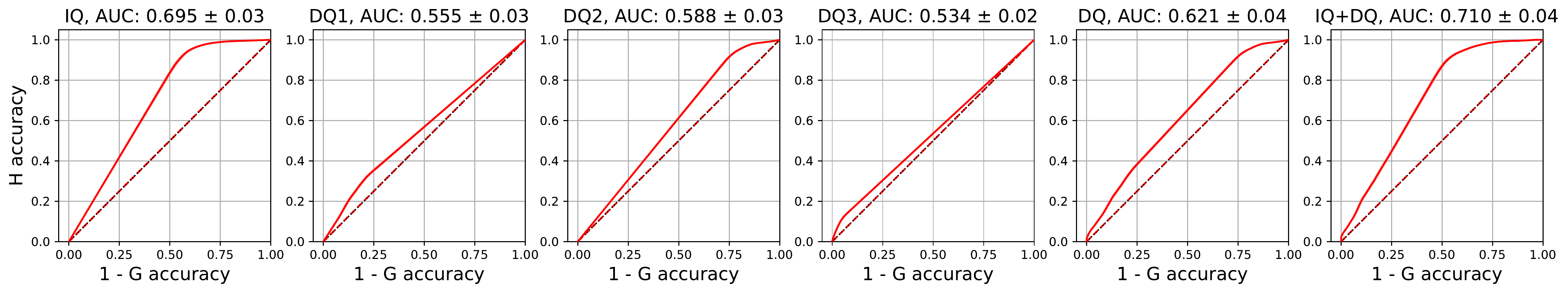}
\caption{GPT-3}
  \label{fig:gpt3_roc}
     \end{subfigure}
        \caption{ROC Curves for the IQ and DQ approaches along with the ensemble approaches
        \label{fig:three_graphs_app}}
\end{figure*}

\begin{figure*}  
\centering  
\newcommand{\widthscale}{0.30}  
\centering  
\includegraphics[width=\widthscale\textwidth]
{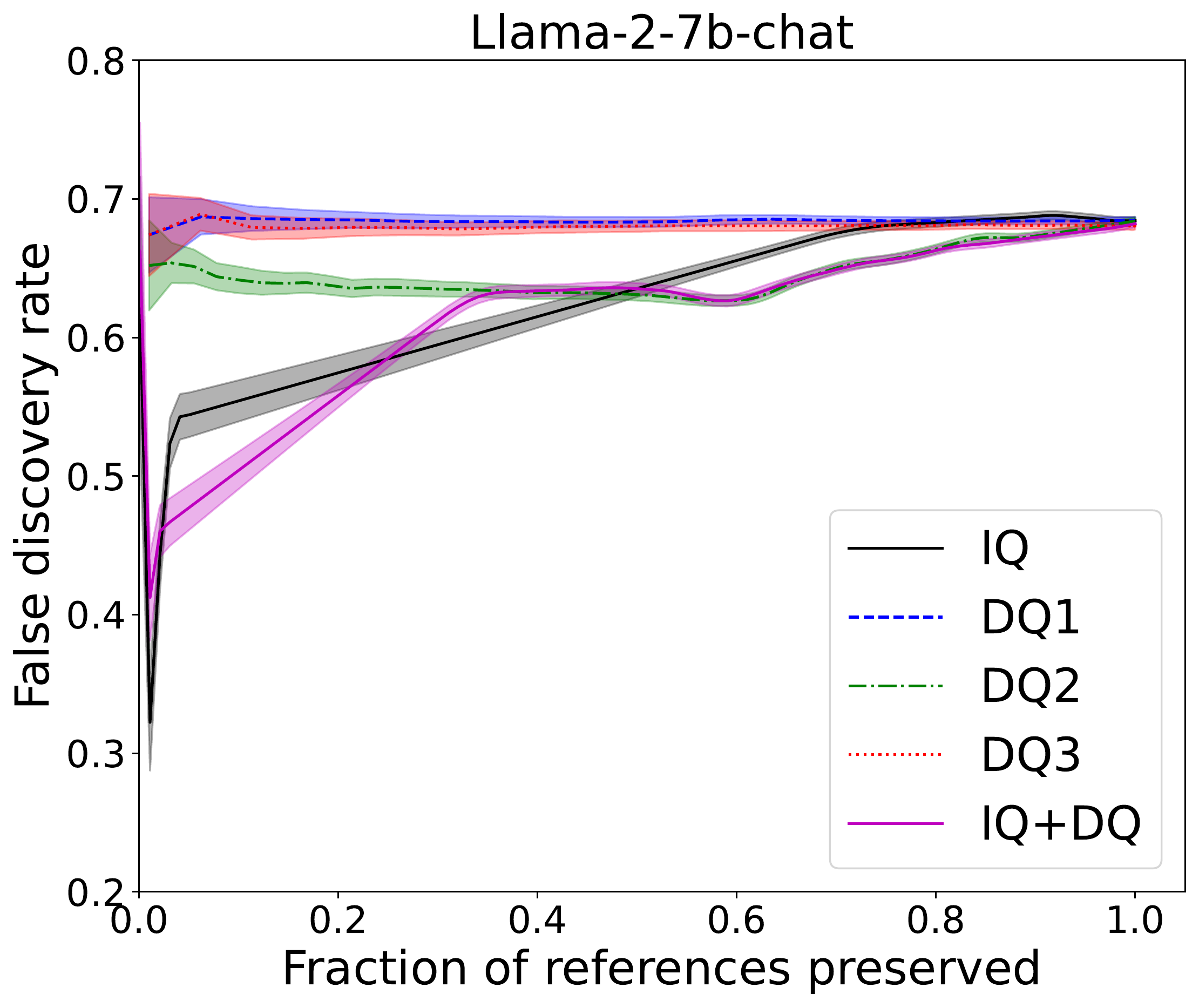}
\includegraphics[width=\widthscale\textwidth]{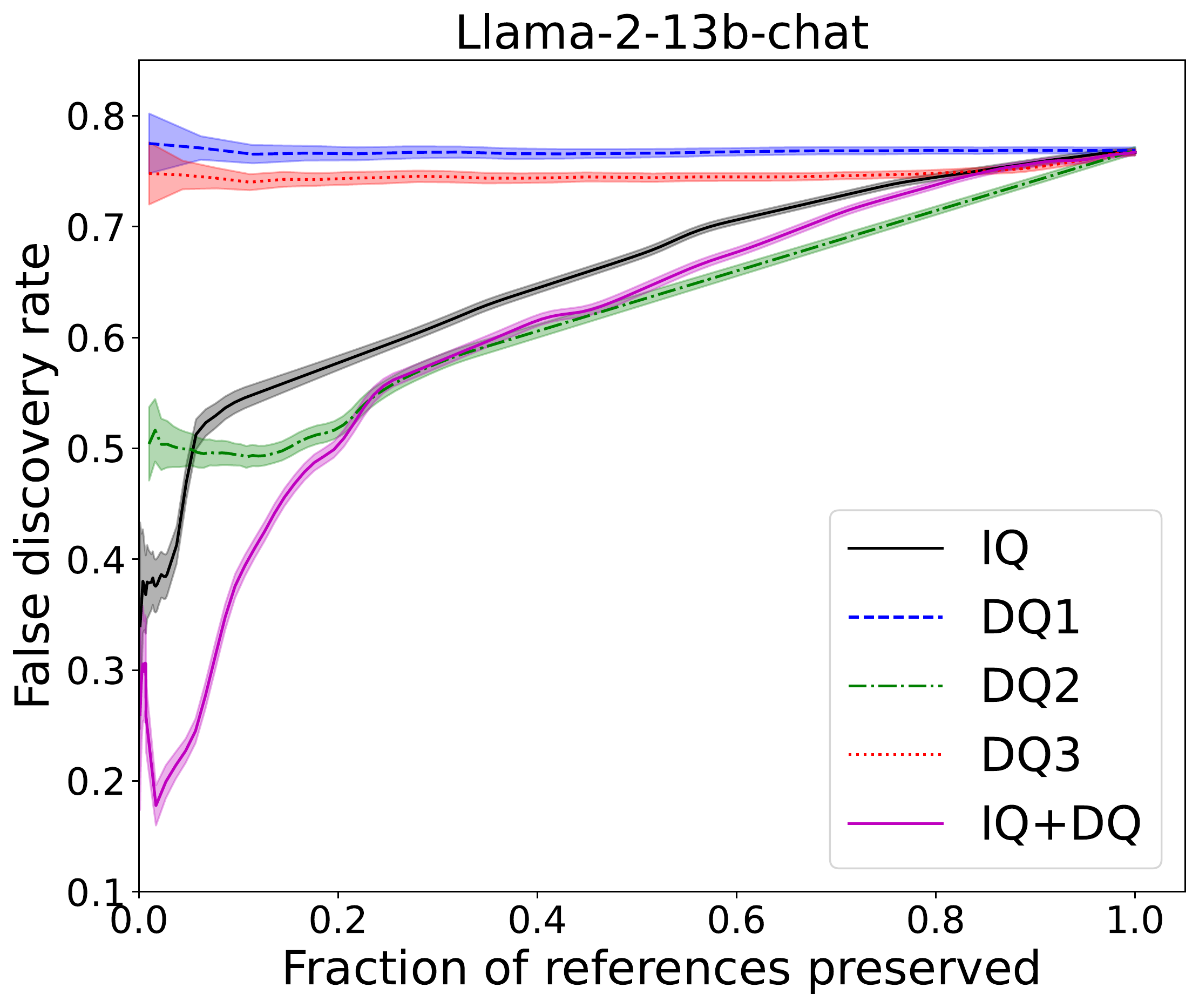}  
\includegraphics[width=\widthscale\textwidth]{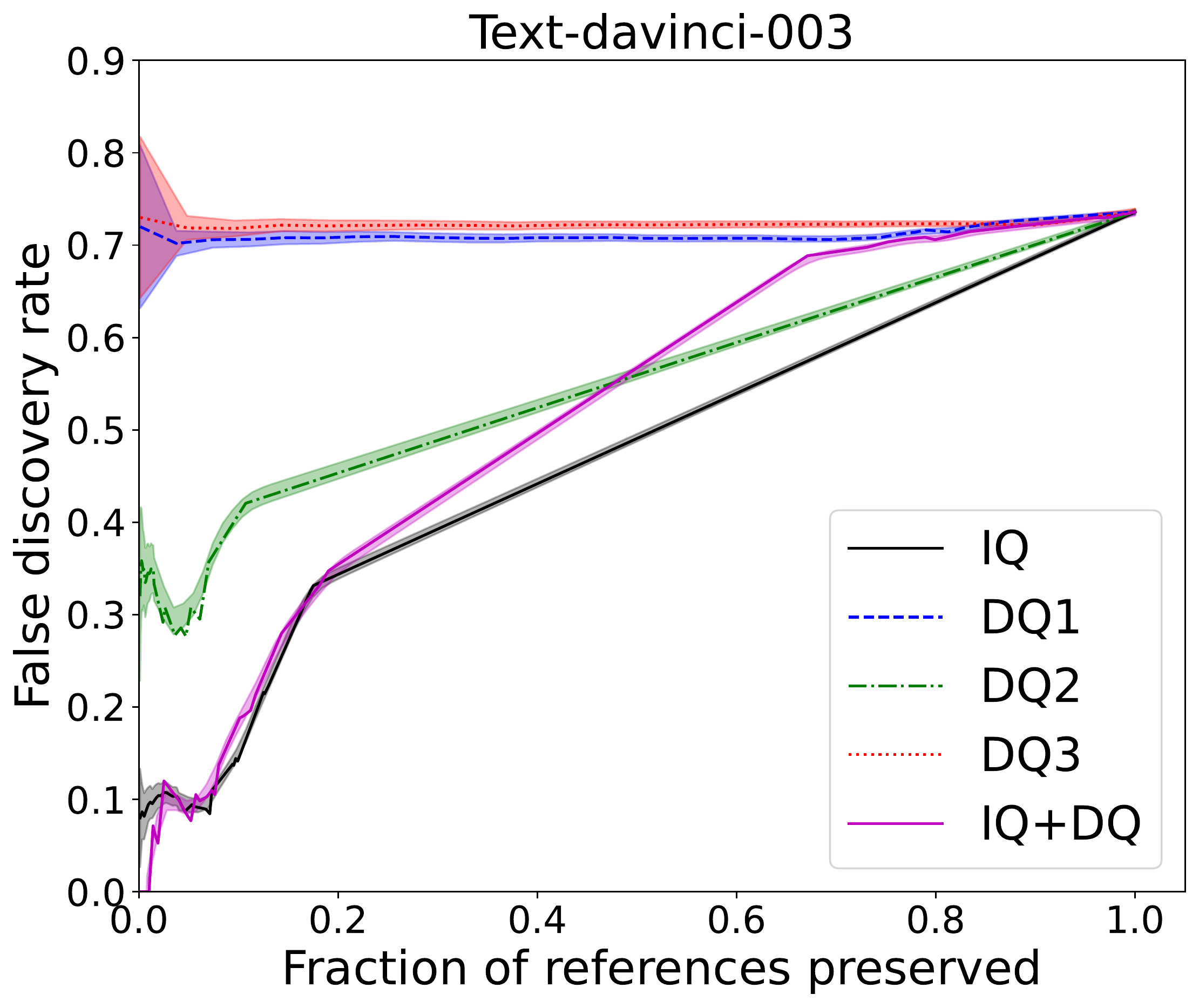}  

\caption{False discovery rate (FDR) vs.~fraction of references preserved for each groundedness filter and LM. The preservation rate indicates the fraction of references preserved when a groundedness filter is applied to the raw generations of a LM. The FDR represents the fraction of preserved references that are actually hallucinations. For unachievable values of the fraction of references preserved (below the minimal fraction achievable by thresholding), we extrapolate each curve by uniformly subsampling references with maximal scores. We compute 95\% confidence intervals from a 100-replicate bootstrap mean $\pm1.96$ times the bootstrap standard
error.}    
\label{fig:fdr_curves_cf_appendix}  
\end{figure*}

We show ROC and FDR metrics for L2-13B, L2-7B and GPT-3 models in \cref{fig:three_graphs_app} and \cref{fig:fdr_curves_cf_appendix} respectively. We find that the procedures are not effective in detecting hallucinations, performing the worst for the L2-7B. Though IQ helps the most for GPT-3, DQ2 approach helps the most for L2-13B and L2-7B. Consistent with our findings of other models, IQ+DQ ensemble approach performs the best.

\section{Licenses and Terms of Use}\label{sec:license}
According to the OpenAI terms of use Sharing and Publication policy,\footnote{\url{https://openai.com/policies/sharing-publication-policy}} %
they ``welcome research publications related to the OpenAI API.'' Following the Bing Search API Legal Information\footnote{\url{https://www.microsoft.com/en-us/bing/apis/legal}}, we do not store the results of the search queries but rather only whether or not there were any results.
According to the ACM,\footnote{\url{https://www.acm.org/publications/class-2012}} ``The full CCS classification tree is freely available for educational and research purposes.'' (This section will be included with any published version of our paper.)

\section{Computation and Cost}\label{sec:compute}
We use OpenAI API for running the experiments on GPT-4, ChatGPT and GPT-3. We show the average tokens consumed for prompt and completion for each of the approaches and data generation per candidate query in  \cref{tab:gpt4_compute,tab:chatgpt_compute,tab:tdavinci_compute}. We estimate the cost based on the pricing details available as of May 2023.\footnote{\url{https://openai.com/pricing}} For GPT-4, around 2.2M tokens were used amounting to roughly \$74 to evaluate all approaches. For ChatGPT, around 2.3M tokens were used amounting to roughly \$5. For GPT-3, around 2.1M tokens were used amounting to roughly \$258.
For Bing Search, we use an S1 instance of the Bing Search API \footnote{\url{https://www.microsoft.com/en-us/bing/apis/pricing}}. We made 3,000 queries in all to this endpoint amounting to \$75. Summing these costs gives a total of \$412. The compute requirements of combining these  
results were negligible. While the exact model sizes and floating point operations are not publicly available for these models, the total cost gives a rough idea on the order of magnitude of computation required in comparison to the hourly cost of, say, a GPU on the Azure platform.  

For running the experiments on Llama-2-chat series, we used a node with $8$ V100 GPUs.

\begin{table*}[th!]
\caption{GPT-4: Average number of tokens consumed}
\centering
\begin{tabular}{cccccc} \toprule
 & \textbf{DS} & \textbf{IQ} & \textbf{DQ1} & \textbf{DQ2} & \textbf{DQ3} \\
\midrule
\textbf{Prompt}     & $40.1$ & $443.4$  & $221.2$  & $299.6$  & $946.1$ \\
\textbf{Completion}     & $64.8$ & $140.1$  & $67.2$  & $12.2$  & $30.3$ \\
\bottomrule
\end{tabular}
\label{tab:gpt4_compute}
\end{table*}
\begin{table*}[th!]
\caption{ChatGPT: Average number of tokens consumed}
\centering
\begin{tabular}{cccccc} \toprule

& \textbf{DS} & \textbf{IQ} & \textbf{DQ1} & \textbf{DQ2} & \textbf{DQ3} \\
\midrule
\textbf{Prompt}     & $40.1$ & $437.3$  & $224.1$  & $302.2$  & $1009.6$ \\
\textbf{Completion}     & $71.8$ & $144.9$  & $28.8$  & $45.5$  & $75.8$ \\
\bottomrule
\end{tabular}
\label{tab:chatgpt_compute}
\end{table*}
\begin{table*}[th!]
\caption{GPT-3: Average number of tokens consumed }
\centering
\begin{tabular}{cccccc} \toprule

&  \textbf{DS} & \textbf{IQ} & \textbf{DQ1} & \textbf{DQ2} & \textbf{DQ3} \\
\midrule
\textbf{Prompt}   &  $39.7$ & $399.53$   & $232.36$  & $332.4$  & $995.1$  \\
\textbf{Completion}  & $68.4$ & $90.6$  & $30.3$ & $21.8$ & $30.4$ \\
\bottomrule
\end{tabular}
\label{tab:tdavinci_compute}
\end{table*}

\section{Examples of Hallucinations and References}\label{sec:examples}

\cref{tab:chatgpt-H,tab:chatgpt-G,tab:gpt4-H,tab:gpt4-G}
 each display a careful inspection of 30 random candidate paper titles classified as H and G as determined by whether the Bing Search API returned any results. A manual search for each suggested title indicated that the vast majority of Hs are in fact hallucinations and the vast majority of Gs are in fact real references. We show the titles classified as H by Bing search along with closest manually discovered match for ChatGPT (\cref{tab:chatgpt-H}) and GPT-4 (\cref{tab:gpt4-H}). We show the titles classified as G by Bing search along with the web links to the matched titles for ChatGPT (\cref{tab:chatgpt-G}) and GPT-4 (\cref{tab:gpt4-G}). We also list the score assigned by the IQ method for all the sampled candidate titles. Interestingly, for both models there was a case in which the IQ method assigned the score of 1 to an H title. These H titles were \textit{Design and Implementation of Digital Libraries: Technological Challenges and Solutions} for ChatGPT (\cref{tab:chatgpt-H}) and \textit{Enterprise Modeling: Tackling Business Challenges with the 4EM Approach} for GPT-4 (\cref{tab:gpt4-H}). In both of these cases, the titles were very similar to the closest manually discovered matched titles - \textit{Design and Implementation of Digital Libraries} and \textit{Enterprise Modeling with 4EM: Perspectives and Method}, respectively.

\begin{table*}
\centering
\caption{Reference titles classified as H (hallucination) by Bing generated from ChatGPT. 30 randomly sampled titles are shown.}
\label{tab:chatgpt-H}
\renewcommand{\arraystretch}{1.2}
\fontsize{9}{11}\selectfont 
\begin{tabular}{p{13cm}l}
\toprule
\textbf{Reference title generated (Closest Match, if found)} &  \textbf{IQ Prob}\\
\midrule
                               Quantum sensing for healthcare (NA) &          0 \\
Challenges and Solutions in Managing Electronic Records in Storage Systems (\href{https://library.osu.edu/osu-records-management/challenges}{Electronic Records Management Challenges}) &          0 \\
Hardware Verification Using Physical Design Techniques (NA) &          0 \\
A Framework for Verifying Recursive Programs with Pointers using Automata over Infinite Trees (\href{https://www.researchgate.net/publication/224088989_Verification_of_recursive_methods_on_tree-like_data_structures}{Verification of recursive methods on tree-like data structures}) &          0 \\
Robust Control for Nonlinear Time-Delay Systems with Faults (\href{https://link.springer.com/book/10.1007/978-981-10-5131-9}{Robust Control for Nonlinear Time-Delay Systems}) &          0 \\
Intelligent Scheduling for Autonomous UAVs using Discrete Artificial Intelligence Planning Techniques (NA) &          0 \\
An Overview of Database Management System Engines for Distributed Computing (NA) &          0 \\
The Aesthetics of Digital Arts and Media (\href{https://direct.mit.edu/books/book/2879/VOICEVocal-Aesthetics-in-Digital-Arts-and-Media}{VOICE: Vocal Aesthetics in Digital Arts and Media}) &          0 \\
Improving Human-Robot Team Performance through Integrated Task Planning and Scheduling in a Complex Environment (\href{https://www.ri.cmu.edu/pub_files/2015/11/The-International-Journal-of-Robotics-Research-2015-Nikolaidis-1711-30.pdf}{Improved human–robot team
performance through cross-training, an
approach inspired by human team
training practices
}) &          0 \\
Web Application Security: From Concept to Practice (\href{https://www.oreilly.com/library/view/web-application-security/9781492053101/}{Web Application Security}) &          0 \\
A 28 nm high-density and low-power standard cell library with half-VDD power-gating cells (NA) &          0 \\
An Acoustic Interface for Touchless Human-Computer Interaction (NA) &          0 \\
Advances in Solid State Lasers Development and Applications: Proceedings of the 42nd Polish Conference on Laser Technology and Applications (\href{https://www.intechopen.com/books/3710}{Advances in Solid State Lasers Development and Applications}) &          0 \\
Designing mobile information systems for healthcare (\href{https://www.researchgate.net/publication/344279303_Design_and_Implementation_of_Mobile-Based_Technology_in_Strengthening_Health_Information_System}{Design and Implementation of Mobile-Based Technology in Strengthening Health Information System}) &          0 \\
Fault-tolerance and Reliability Techniques for Dependable Distributed Systems (\href{https://www.researchgate.net/publication/340902878_RELIABILITY_AND_REPLICATION_TECHNIQUES_FOR_IMPROVED_FAULT_TOLERANCE_IN_DISTRIBUTED_SYSTEMS}{Reliability and Replication Techniques for Improved Fault Tolerance in Distributed Systems}) &          0 \\
Cyber-physical systems: A Survey and Future Research Directions on Sensor and Actuator Integration (\href{https://www.researchgate.net/publication/289713317_Cyber-physical_systems_A_survey}{Cyber-physical systems: A survey}) &          0 \\
Performance evaluation of wireless sensor networks using network simulator-3 (NA) &          0 \\
Communication-Based Design for VLSI Circuits and Systems (NA) &          0 \\
Digital Media: The Intersection of Art and Technology (NA) &          0 \\
Toward a tool-supported software evolution methodology (NA) &          0 \\
Performance evaluation of temperature-aware routing protocols in wireless sensor networks (\href{https://www.researchgate.net/publication/43121788_Performance_Evaluation_of_Routing_Protocols_in_Wireless_Sensor_Networks}{Performance Evaluation of Routing Protocols in Wireless Sensor Networks})  &          0 \\
Computer-managed instruction and student learning outcomes: a meta-analysis (\href{https://www.researchgate.net/publication/234647833_Effects_of_Computer-Assisted_Instruction_on_Cognitive_Outcomes_A_Meta-Analysis#:~:text=A\%20meta-analysis\%20was\%20performed\%20to\%20synthesize\%20existing\%20research,quantitative\%20data\%20were\%20transformed\%20into\%20Effect\%20Size\%20\%28ES\%29.}{Effects of Computer-Assisted Instruction on Cognitive Outcomes: A Meta-Analysis}) &          0 \\

An Empirical Analysis of Enterprise Resource Planning (ERP) Systems Implementation in Service Organizations in Jordan (\href{https://www.researchgate.net/publication/260870257_Contributions_of_ERP_Systems_in_Jordan}{Contributions of ERP Systems in Jordan}) &          0 \\

Optimization of production planning in consumer products industry (\href{https://ltplabs.com/success-cases/optimizing-production-planning-at-a-consumer-goods-company/}{Optimizing production planning at a consumer goods company}) &          0.01 \\
Efficient Text Document Retrieval Using an Inverted Index with Cache Enhancement (NA) &          0.11 \\
Service OAM in Carrier Ethernet Networks &          0.13 \\
Introduction to Logic: Abstraction in Contemporary Logic (\href{https://dorshon.com/wp-content/uploads/2018/03/Introduction-to-Logic.pdf}{Introduction to Logic}) &          0.17 \\
Query Processing and Optimization for Information Retrieval Systems (\href{https://redirect.cs.umbc.edu/courses/graduate/676/SP2021/termpapers/CMSC476676-TermPaperBirmalShivani.pdf}{Query Optimization in Information Retrieval}) &          0.33 \\
Cross-Platform Verification of Web Applications (\href{https://dl.acm.org/doi/10.1145/2610384.2610409}{Cross-platform feature matching for web applications}) &          0.33 \\
Design and Implementation of Digital Libraries: Technological Challenges and Solutions (\href{https://www.academia.edu/23683514/Chapter_18_DESIGN_AND_IMPLEMENTATION_OF_DIGITAL_LIBRARIES}{Design and Implementation of Digital Libraries}) &          1 \\
\bottomrule
\end{tabular}
\end{table*}

\begin{table*}
\centering
\caption{Reference titles classified as G (grounded) by Bing, generated from ChatGPT. 30 randomly sampled titles are shown.}
\label{tab:chatgpt-G}
\renewcommand{\arraystretch}{1.2}
\begin{tabular}{p{12cm}lp{10cm}}
\toprule
\textbf{Reference title generated (Matched title)} &  \textbf{IQ Prob}\\
\midrule
JavaScript: The Good Parts (\href{https://www.oreilly.com/library/view/javascript-the-good/9780596517748/}{exact match}) &          1 \\
Essentials of Management Information Systems (\href{https://books.google.co.in/books/about/Essentials_of_Management_Information_Sys.html?id=TyAvPwAACAAJ&redir_esc=y}{exact match}) &          1 \\
Visualization Analysis and Design (\href{https://www.taylorfrancis.com/books/mono/10.1201/b17511/visualization-analysis-design-tamara-munzner}{exact match}) &          1 \\
Forecasting: Methods and Applications 
(\href{https://www.google.com/books/edition/FORECASTING_METHODS_AND_APPLICATIONS_3RD/nxt0CgAAQBAJ?hl=en}{exact match}) &          1 \\
Python for Data Analysis (\href{https://wesmckinney.com/book/}{exact match}) &          1 \\
Introduction to Parallel Algorithms and Architectures: Arrays Trees Hypercubes (\href{https://dl.acm.org/doi/pdf/10.1145/141914.990672}{exact match}) &          1 \\
Linear logic and its applications (\href{https://citeseerx.ist.psu.edu/document?repid=rep1&type=pdf&doi=740ffd6d00518b11976cf6eed790246923de737f}{Temporal Linear Logic and Its
Applications}) &          1 \\
Coding and Information Theory (\href{https://link.springer.com/book/9780387978123}{exact match}) &          1 \\
Introduction to Electric Circuits (\href{https://archive.org/details/IntroductionToElectricCircuits9thEd}{exact match}) &          1 \\
Concurrent Programming in Java: Design Principles and Patterns (\href{https://leon-wtf.github.io/doc/concurrent\%20programming\%20in\%20java\%20design\%20principles\%20and\%20pattern.pdf}{exact match}) &          1 \\
Cross-Platform GUI Programming with wxWidgets (\href{https://www.wxwidgets.org/docs/book/}{exact match}) &          1 \\
Embedded Computing and Mechatronics with the PIC32 Microcontroller (\href{https://www.sciencedirect.com/book/9780124201651/embedded-computing-and-mechatronics-with-the-pic32-microcontroller}{exact match}) &          0.87 \\
Quantum entanglement for secure communication (\href{https://www.zdnet.com/article/quantum-entanglement-breakthrough-could-boost-encryption-secure-communications/}{Quantum entanglement breakthrough could boost encryption, secure communications}) &          0.78 \\
An Introduction to Topology and its Applications (\href{https://repository.nie.edu.sg/bitstream/10497/14313/1/CAME2012_a.pdf}{An introduction to topology and its applications: A new approach}) &          0.67 \\
SQL Server Query Performance Tuning (\href{https://books.google.co.in/books?hl=en&lr=&id=C65qBAAAQBAJ&oi=fnd&pg=PR27&dq=\%22SQL+Server+Query+Performance+Tuning\%22+&ots=El1cheYbL4&sig=CeSyKIh9d2lEXwqct8SBppwCa_0&redir_esc=y#v=onepage&q=\%22SQL\%20Server\%20Query\%20Performance\%20Tuning\%22&f=false}{exact match}) &          0.67 \\
WCAG 2.1: Web Content Accessibility Guidelines (\href{https://www.w3.org/TR/WCAG21/}{exact match}) &          0.61 \\
Session Announcement Protocol (SAP) (\href{https://www.cl.cam.ac.uk/~jac22/books/mm/book/node184.html}{exact match}) &          0.5 \\
Introduction to Atmospheric Chemistry (\href{https://acmg.seas.harvard.edu/education/introduction-atmospheric-chemistry}{exact match}) &          0.33 \\
Data modeling and database design: Using access to build a database (\href{https://rollmeup.willienelson.com/g/pdf/H3H1E2/data-modeling-and-database-design-using-access-to-build-a-database_pdf}{exact match}) &          0.33 \\
Introductory Digital Electronics: From Truth Tables to Microprocessors (\href{https://annas-archive.org/md5/1966a08980103ff534ade06bd1af65ec}{exact match}) &          0.33 \\
Trust Management: First International Conference, iTrust 2003, Heraklion, Crete, Greece (\href{https://researchprofiles.canberra.edu.au/en/publications/trust-management-first-international-conference-itrust-2003-herak}{exact match}) &          0.25 \\
Random geometric graphs (\href{https://academic.oup.com/book/9064}{exact match}) &          0.08 \\
Statistical Inference: An Integrated Approach (\href{https://www.routledge.com/Statistical-Inference-An-Integrated-Approach-Second-Edition/Migon-Gamerman-Louzada/p/book/9781439878804}{exact match}) &          0 \\
Network Service Assurance (\href{https://www.hcltech.com/engineering-rd-services/network-service-assurance}{exact match}) &          0 \\
Higher Order Equational Logic Programming (\href{https://dl.acm.org/doi/10.1145/174675.177889#:~:text=\%20Higher-order\%20equational\%20logic\%20programming\%20is\%20a\%20paradigm,subclass\%20of\%20simply\%20typed\%20\%CE\%BB-terms\%2C\%20called\%20higher-order\%20patterns.}{exact match}) &          0 \\
Network Mobility Route Optimization Requirements (\href{https://www.semanticscholar.org/paper/Network-Mobility-Route-Optimization-Requirements-in-Eddy-Ivancic/6273756bafe55b9595d1706835ddfc6f5b83d09d}{Network Mobility Route Optimization Requirements for Operational Use in Aeronautics and Space Exploration Mobile Networks}) &          0 \\
Thermal management of electric vehicle battery systems (\href{https://onlinelibrary.wiley.com/doi/book/10.1002/9781118900239}{exact match}) &          0 \\
Handbook of Imaging Materials (\href{https://www.taylorfrancis.com/books/edit/10.1201/9781315214597/handbook-imaging-materials-arthur-diamond}{exact match}) &          0 \\
The Secure Online Business Handbook: E-commerce, IT Functionality and Business Continuity (\href{https://books.google.co.in/books?hl=en&lr=&id=L-MxmGIs1n0C&oi=fnd&pg=PR11&dq=\%22The+Secure+Online+Business+Handbook:+E-commerce,+IT+Functionality+and+Business+Continuity\%22+&ots=stN79AbBeU&sig=PptSBuD-E9rdHIh9h4BXoCDUDy4&redir_esc=y#v=onepage&q=\%22The\%20Secure\%20Online\%20Business\%20Handbook\%3A\%20E-commerce\%2C\%20IT\%20Functionality\%20and\%20Business\%20Continuity\%22&f=false}{exact match}) &          0 \\
Advanced Logic Synthesis (\href{https://link.springer.com/book/10.1007/978-3-319-67295-3}{exact match}) &          0 \\
\bottomrule
\end{tabular}
\end{table*}

\begin{table*}
\centering
\caption{Reference titles classified as H (hallucination) by Bing generated from GPT-4. 30 randomly sampled titles are shown.}
\label{tab:gpt4-H}
\renewcommand{\arraystretch}{1.2}
\fontsize{10}{11}\selectfont 
\begin{tabular}{p{14cm}l}

\toprule
\textbf{Reference title generated (Closest Match, if found)} &  \textbf{IQ Prob}\\
\midrule
Privacy-Preserving Attribute-Based Access Control in Cloud Computing (\href{https://link.springer.com/article/10.1007/s10207-021-00565-4}{Accountable privacy preserving attribute-based access control for cloud services enforced using blockchain}) &          0 \\

Policy Measures for Combating Online Privacy Issues (NA) &          0 \\

Storage Security: Protecting Sanitized Data Attestation (NA) &          0 \\

Design of Scalable Parallel Algorithms for Graph Problems (NA) &          0 \\

Very Large Scale Integration (VLSI) Design with Standard Cells: Layout Design and Performance Analysis (NA) &          0 \\

Object-Oriented Modeling and Simulation of Complex Systems (\href{https://citeseerx.ist.psu.edu/document?repid=rep1&type=pdf&doi=2e0e69255a516478dd8824f29e55496492eed932}{Modelling and simulation of
complex systems}) &          0 \\

Overview of Electronic Design Automation (EDA) Tools \& Methodologies (\href{https://link.springer.com/book/10.1007/978-0-387-73543-6}{The Electronic Design Automation Handbook}) &          0 \\

Printers and Modern Storage Solutions: The Role of the Cloud and Mobile Devices (NA) &          0 \\

Algebraic Algorithms and Symbolic Analysis Techniques in Computer Algebra Systems (\href{https://dl.acm.org/doi/abs/10.5555/153180}{Computer algebra systems and algorithms for algebraic computation}) &          0 \\

Measuring Software Performance in Cross-platform Mobile Applications (NA) &          0 \\

A Comparative Study of OAM Protocols in Ethernet Networks (\href{https://link.springer.com/chapter/10.1007/978-3-642-21484-4_14}{Carrier Ethernet OAM: an overview and comparison to IP OAM}) &          0 \\

Best Practices in Board- and System-level Hardware Test Development (NA) &          0 \\

Algorithms for Symbolic and Algebraic Computations in Science and Engineering (NA) &          0 \\

Cryptography and Secure E-Commerce Transactions: Methods, Frameworks, and Best Practices (NA) &          0 \\

Quantum Computing: A Primer for Understanding and Implementation (
  \href{https://link.springer.com/content/pdf/10.1007/978-3-030-19066-8.pdf}{A primer on quantum computing}
) &          0 \\

Understanding Network Management: Concepts, Standards, and Models (\href{https://books.google.com/books?hl=en&lr=&id=VGDMIIfL6XcC&oi=fnd&pg=PR23&dq=Understanding+Network+Management:+Concepts,+Standards,+and+Models&ots=mdIIcM6mb4&sig=mddEdLY35vOxZVwlbaOVX1KbHY4}{Network management: principles and practice}) &          0 \\

Assessing network reliability: An analytical approach based on graph entropy (NA) &          0 \\

Language Models and their Applications to Information Retrieval (\href{https://nlp.stanford.edu/IR-book/html/htmledition/language-models-for-information-retrieval-1.html}{Language models for information retrieval}) &          0 \\

Automated Support for Legacy Software Maintenance and Evolution (NA) &          0 \\

In-Network Traffic Processing: Advancements and Perspectives (NA) &          0 \\
Intellectual Property Law and Policy in the Digital Economy (\href{https://heinonline.org/HOL/LandingPage?handle=hein.journals/hulr44&div=37&id=&page=}{Intellectual Property Law and Policy in the Digital Economy})  &          0 \\

The Art and Science of Survey Research: A Guide to Best Practices (\href{https://www.researchgate.net/publication/291363048_The_Art_and_Science_of_Reviewing_and_Writing_Survey_Research}{The Art and Science of Reviewing (and Writing) Survey Research}) &          0 \\

Review of Network Mobility Protocols: Solutions and Challenges (\href{https://www.researchgate.net/publication/264087172_A_Review_of_Network_Mobility_Protocols_for_Fully_Electrical_Vehicles_Services}{A Review of Network Mobility Protocols for Fully Electrical Vehicles Services}) &          0 \\

Program Semantics, Higher-Order Types, and Step Counting (NA) &          0 \\

Network Services: Management Strategies and Techniques (NA) &          0 \\

Machine Learning-Based Power Estimation and Management in Energy Harvesting Systems (NA) &          0 \\
The Evolution of Distance Education: Historical and Theoretical Perspectives (\href{http://dlkkhsou.inflibnet.ac.in/bitstream/123456789/17/9/09_chapter3.pdf}{Distance Education: Historical Perspective}) &          0.17 \\

The Economics of VLSI Manufacturing: A Cost Analysis Approach (NA) &          0.5 \\

Digital Decisions: The Intersection of e-Government and American Federalism (NA) &          0.78 \\
Enterprise Modeling: Tackling Business Challenges with the 4EM Approach (\href{https://link.springer.com/chapter/10.1007/978-3-030-93547-4_5}{Enterprise Modeling with 4EM: Perspectives and Method}) &          1 \\
\bottomrule
\end{tabular}
\end{table*}

\begin{table*}
\centering
\caption{Reference titles classified as G (grounded) by Bing generated from GPT-4. 30 randomly sampled titles are shown.}
\label{tab:gpt4-G}
\renewcommand{\arraystretch}{1.3}
\begin{tabular}{p{13cm}l}
\toprule
\textbf{Reference title generated (Matched title)} &  \textbf{IQ Prob}\\
\midrule
Art and Electronic Media (\href{https://books.google.co.in/books/about/Art_and_Electronic_Media.html?id=CeTsbAlzEN8C&redir_esc=y}{exact match}) &          1 \\

Network+ Guide to Networks (\href{https://dl.acm.org/doi/book/10.5555/3235279}{exact match}) &          1 \\

Handbook of Automated Reasoning (\href{https://dl.acm.org/doi/10.5555/581809}{exact match}) &          1 \\

System Dynamics: Modeling, Simulation, and Control of Mechatronic Systems (\href{https://www.wiley.com/en-us/System+Dynamics\%3A+Modeling\%2C+Simulation\%2C+and+Control+of+Mechatronic+Systems\%2C+5th+Edition-p-9780470889084}{exact match}) &          1 \\

Information Visualization: Perception for Design (\href{https://books.google.co.in/books/about/Information_Visualization.html?id=qFmS95vf6H8C&redir_esc=y#:~:text=Information\%20Visualization\%3A\%20Perception\%20for\%20Design\%20is\%20a\%20comprehensive,is\%20a\%20super-computer\%20for\%20finding\%20patterns\%20in\%20information.}{exact match}) &          1 \\

The Emperor's New Mind: Concerning Computers, Minds and the Laws of Physics (\href{https://oceanofpdf.com/authors/roger-penrose/pdf-epub-the-emperors-new-mind-concerning-computers-minds-and-the-laws-of-physics-download/}{exact match}) &          1 \\

Computer Networks: A Systems Approach (\href{https://open.umn.edu/opentextbooks/textbooks/771}{exact match}) &          1 \\

DNS and BIND: Help for System Administrators (\href{https://books.google.co.in/books/about/DNS_and_BIND.html?id=HggtWI1ShvMC&redir_esc=y}{exact match}) &          1 \\

Introduction to Modern Cryptography (\href{https://www.cs.ucdavis.edu/~rogaway/classes/227/fall01/book/main.pdf}{exact match}) &          1 \\

Beyond Software Architecture: Creating and Sustaining Winning Solutions (\href{https://ieeexplore.ieee.org/document/1259275}{exact match}) &          1 \\

Practical Byzantine Fault Tolerance and Proactive Recovery (\href{https://dl.acm.org/doi/10.1145/571637.571640}{exact match}) &          1 \\

Real-Time Systems: Scheduling, Analysis, and Verification (\href{https://www.semanticscholar.org/paper/Real-time-systems-scheduling\%2C-analysis\%2C-and-Cheng/0dcd34a9f91d0bd63541288c1cbdfefff1c71e2f}{exact match}) &          1 \\

Computational Complexity: A Modern Approach (\href{https://theory.cs.princeton.edu/complexity/book.pdf}{exact match}) &          1 \\

The Foundations of Cryptography: Volume 1, Basic Techniques (\href{https://www.researchgate.net/publication/220688776_The_Foundations_of_Cryptography_-_Volume_1_Basic_Techniques}{exact match}) &          1 \\

Digital Library Use: Social Practice in Design and Evaluation (\href{https://direct.mit.edu/books/book/3828/Digital-Library-UseSocial-Practice-in-Design-and}{exact match}) &          1 \\

Transactional Information Systems: Theory, Algorithms, and the Practice of Concurrency Control and Recovery (\href{https://dl.acm.org/doi/book/10.5555/378243}{exact match}) &          1 \\

Database System Concepts (\href{https://www.bing.com/search?q=\%22Database+System+Concepts\%22&cvid=7aa84d9877094c4e9383b55e48ba1581&aqs=edge..69i57j0l8j69i11004.4459j0j4&FORM=ANAB01&PC=U531}{exact match}) &          1 \\

Pattern Recognition and Machine Learning (\href{https://link.springer.com/book/9780387310732}{exact match}) &          1 \\

File System Forensic Analysis (\href{https://dl.acm.org/doi/book/10.5555/1051914}{exact match}) &          1 \\

The Archaeology of Science: Studying the Creation of Useful Knowledge (\href{https://books.google.co.in/books/about/The_Archaeology_of_Science.html?id=qcNDAAAAQBAJ&redir_esc=y}{exact match}) &          0.78 \\

Web Data Mining: Exploring Hyperlinks, Contents, and Usage Data (\href{https://link.springer.com/book/10.1007/978-3-642-19460-3}{exact match}) &          0.67 \\

Electronic Design Automation for Integrated Circuits Handbook (\href{https://www.routledge.com/Electronic-Design-Automation-for-Integrated-Circuits-Handbook-Second-Edition/Lavagno-Markov-Martin-Scheffer/p/book/9781032339986}{exact match}) &          0.47 \\

Modern VLSI Design: IP-Based Design (\href{http://www.csit-sun.pub.ro/courses/vlsi/Modern_VLSI_Design.pdf}{exact match}) &          0.39 \\

Computational Complexity and Statistical Physics (\href{https://academic.oup.com/book/41797}{exact match}) &          0.33 \\

Probabilistic Methods for Algorithmic Discrete Mathematics (\href{https://link.springer.com/book/10.1007/978-3-662-12788-9}{exact match}) &          0.33 \\

Digital Rights Management: Protecting and Monetizing Content (\href{https://www.routledge.com/Digital-Rights-Management-Protecting-and-Monetizing-Content/Tassel/p/book/9780240807225}{exact match}) &          0.08 \\

Deep Learning for Computer Vision: A Brief Review (\href{https://dl.acm.org/doi/abs/10.1155/2018/7068349}{exact match}) &          0.08 \\

Random Geometric Graphs and Applications (\href{https://www.proquest.com/openview/c5c66e3857cfa928a07b6bd08506ad56/1?pq-origsite=gscholar&cbl=18750&diss=y}{exact match}) &          0.07 \\

Concurrent Separation Logic for Pipelined Parallelization (\href{https://www.cs.princeton.edu/~appel/papers/cslchannels.pdf}{exact match}) &          0 \\

High-Level Synthesis for Real-time Digital Signal Processing (\href{https://link.springer.com/book/10.1007/978-1-4757-2222-2}{exact match}) &          0 \\
\bottomrule
\end{tabular}
\end{table*}

\end{document}